\documentclass[10pt,twocolumn,letterpaper]{article}

\usepackage[pagenumbers]{iccv} 
\definecolor{iccvblue}{rgb}{0.21,0.49,0.74}
\usepackage[pagebackref,breaklinks,colorlinks,allcolors=iccvblue]{hyperref}

\usepackage[linesnumbered,ruled,vlined]{algorithm2e}
\usepackage{courier}
\def\paperID{11370} 
\def\confName{ICCV}
\def\confYear{2025}
\def\Method{SSD}

\usepackage{pifont}
\newcommand{\cmark}{\ding{51}}%
\newcommand{\xmark}{\ding{55}}%

\usepackage{subcaption}

\usepackage{amsthm}
\newtheoremstyle{upright}%
  {3pt} 
  {3pt} 
  {}    
  {}    
  {\bfseries} 
  {.}   
  { }   
  {}    

\theoremstyle{upright}
\newtheorem{definition}{Definition}

\newcommand{\circled}[1]{\textcircled{\raisebox{-0.9pt}{#1}}}

\title{Soft Separation and Distillation: Toward Global Uniformity in \\ Federated Unsupervised Learning}

\author{
Hung-Chieh Fang$^{1,}$\thanks{~Project was partially completed during the author's visit to the CUHK} \quad
Hsuan-Tien Lin$^{1}$ \quad
Irwin King$^{2}$ \quad
Yifei Zhang$^{2,}$\thanks{~Corresponding Author.} \\ [5mm]
$^1$National Taiwan University $^2$The Chinese University of Hong Kong
}

\begin{document}
\maketitle

\begin{abstract}
Federated Unsupervised Learning (FUL) aims to learn expressive representations in federated and self-supervised settings. The quality of representations learned in FUL is usually determined by uniformity, a measure of how uniformly representations are distributed in the embedding space. However, existing solutions perform well in achieving intra-client (local) uniformity for local models while failing to achieve inter-client (global) uniformity after aggregation due to non-IID data distributions and the decentralized nature of FUL. To address this issue, we propose Soft Separation and Distillation (\Method{}), a novel approach that preserves inter-client uniformity by encouraging client representations to spread toward different directions. This design reduces interference during client model aggregation, thereby improving global uniformity while preserving local representation expressiveness. We further enhance this effect by introducing a projector distillation module to address the discrepancy between loss optimization and representation quality. We evaluate \Method{} in both cross-silo and cross-device federated settings, demonstrating consistent improvements in representation quality and task performance across various training scenarios. Our results highlight the importance of inter-client uniformity in FUL and establish \Method{} as an effective solution to this challenge. Project page: \href{https://ssd-uniformity.github.io/}{\small\texttt{https://ssd-uniformity.github.io/}}.
\end{abstract}

\section{Introduction}
\label{sec:intro}
\begin{figure}
    \centering
    \includegraphics[width=0.8\linewidth]{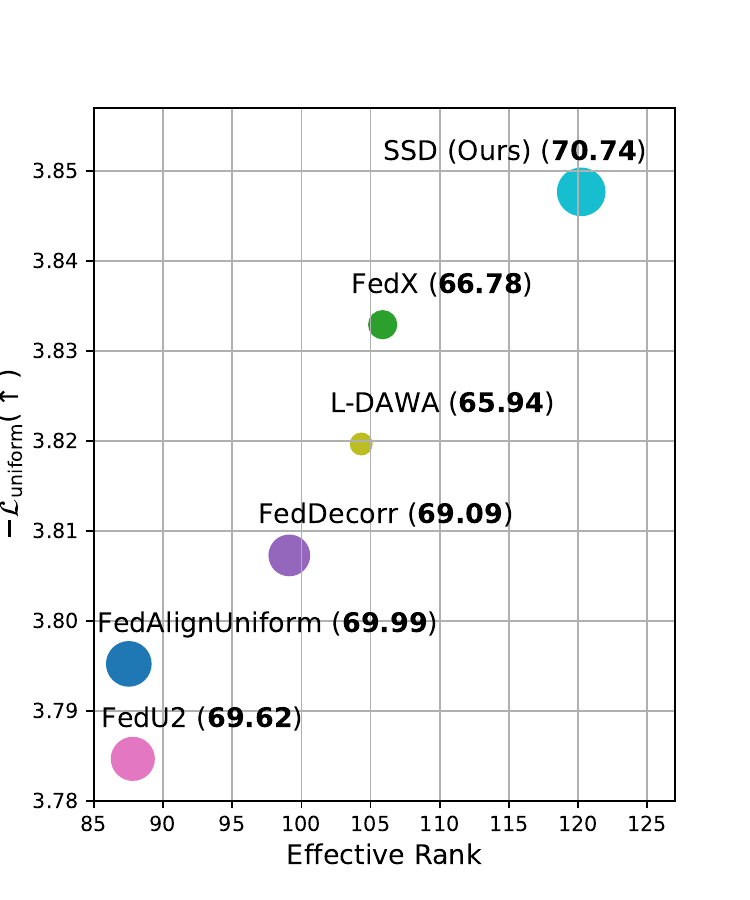}
     \vspace{-10px}
    \caption{\textbf{Comparison with state-of-the-art methods.} Our SSD achieves the \emph{highest} performance (\textbf{marked in brackets}) and the best representation quality, measured by uniformity~\cite{wang2020alignuniform} and effective rank~\cite{roy2007effrank}, among existing FUL approaches.
    \label{fig:effrank_unifority}}
    \vspace{-15px}
\end{figure}
Deep learning has achieved remarkable success across a wide spectrum of applications, from computer vision and natural language processing to speech recognition and reinforcement learning~\cite{he2016resnet, vaswani2017attention, baevski2020wav2vec, levine2016e2epolicy}. This progress can be largely attributed to the availability of massive labeled datasets, particularly since the emergence of ImageNet ~\cite{deng2009imagenet}, which has enabled the training of increasingly powerful neural networks. However, the practical deployment of deep learning in real-world scenarios faces two critical challenges.
First, data in real-world applications is often non-Independent and Identically Distributed (non-IID) and cannot be freely shared due to privacy concerns, regulatory restrictions, or proprietary constraints. This data is typically generated across diverse sources, including user devices, healthcare institutions, and industrial settings. The distributed nature and sharing limitations of such data make centralized training approaches infeasible, leading to the development of Federated Learning (FL) ~\cite{mcmahan2017fedavg, li2020flchallenge, kairouz2021fedadvances}  as a privacy-preserving distributed learning paradigm. Second, a vast proportion of available data remains unlabeled, necessitating effective methods for unsupervised representation learning ~\cite{chen2020simclr, he2022mae, chen2021simsiam, grill2020byol}. These methods aim to learn meaningful feature representations without relying on explicit labels, enabling models to capture underlying data structures and semantic relationships.

Federated Unsupervised Learning (FUL) emerges at the intersection of these two challenges, aiming to learn expressive representations in settings where data is both distributed and unlabeled. In FUL, the quality of learned representations is critically determined by two properties: alignment and uniformity~\cite{wang2020alignuniform}. Alignment measures how close similar data points are positioned in the representation space, relating to the model's ability to group semantically similar objects. Uniformity quantifies how evenly representations are distributed across the unit hypersphere, essentially measuring the entropy of the representation distribution. High uniformity prevents representation collapse and ensures that the learned features effectively utilize the available embedding dimensions. 

However, existing FUL methods face a significant challenge that remains inadequately addressed. While current approaches successfully achieve good intra-client (local) uniformity for representations within each client, they struggle to maintain inter-client (global) uniformity after model aggregation. This limitation stems from two key factors: (1) the non-IID distribution of data across clients, which naturally leads to divergent updates, and (2) the decentralized nature of FL, where the server lacks direct access to raw data, preventing the application of explicit uniformity constraints across clients. 

Most existing approaches have focused primarily on two aspects of this challenge. One line of research builds upon the framework established by FedProx~\cite{li2020fedprox}, introducing proximal terms to constrain local updates within a bounded neighborhood of the global model. For instance, FedU~\cite{zhuang2021fedu} and FedEMA~\cite{zhuang2022fedema} dynamically adjust aggregation weights based on inter-client model divergence, while FedX~\cite{han2022fedx} incorporates a global relational loss to align pairwise sample relationships across clients. These methods aim to maintain global model consistency but do not explicitly address representation uniformity.

Another research direction tackles the issue of dimensional collapse~\cite{jing2022dc} caused by lower uniformity in local representations. FedDecorr~\cite{shi2023feddecorr} demonstrates that embeddings of local clients are often less uniformly distributed and mitigates this issue by decorrelating local features. Similarly, FedU2~\cite{liao2024fedu2} regularizes local updates to approximate a spherical Gaussian distribution, encouraging isotropic feature spaces. While these approaches enhance local uniformity, the improvements at the client level do not inherently translate to better global uniformity during model aggregation. This leads to a crucial question:
\begin{center}
    \emph{How can we effectively improve inter-client uniformity in Federated Unsupervised Learning?}
\end{center}

\begin{figure*}
    \centering
    \includegraphics[width=0.8\linewidth, trim={0.6cm 0 0.6cm 0},clip]{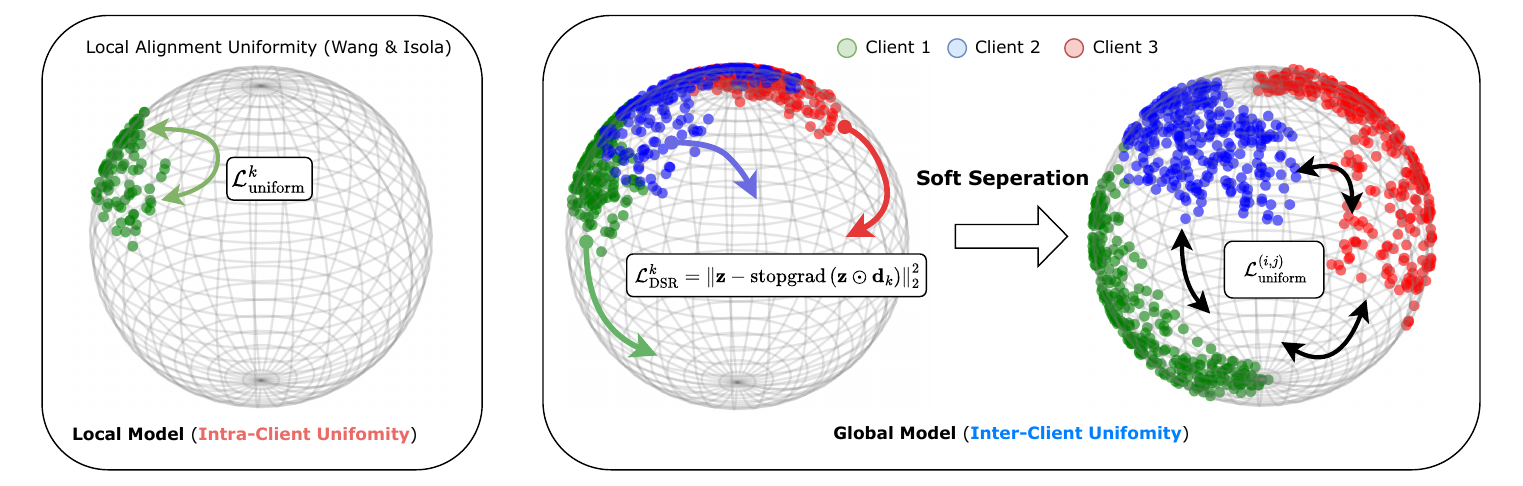}
    \caption{\textbf{Illustration of Intra-Client and Inter-Client Uniformity with Soft Separation.} Intra-client uniformity ensures that representations within each client are well-distributed, while inter-client uniformity promotes global representation consistency across clients. Our proposed Soft Separation method encourages each client's representations to spread in distinct directions, mitigating interference during model aggregation and improving global uniformity.}
    \label{fig:overview}
\end{figure*}

To address this challenge, we propose Soft Separation and Distillation (SSD), a novel method that enhances inter-client uniformity without compromising local representation expressiveness. Our approach employs a dimension-scaled regularization strategy that softly separates each client's feature space, encouraging client representations to spread toward different directions. As illustrated in Figure ~\ref{fig:overview}, this technique reduces interference during client model aggregation, thereby improving global uniformity without imposing rigid boundaries that could distort the underlying feature distributions.

Furthermore, we empirically observe that the regularization effect of SSD may not always effectively transfer from loss optimization to the representation space due to the presence of a projector between the loss function and the encoder. While removing the projector might seem like a straightforward solution, doing so often degrades performance because the projector plays a crucial role in separating optimization objectives from feature representations ~\cite{chen2020simclr, gupta2022projectionhead, xue2024projectionhead}. To bridge this gap, we introduce a projector distillation module that minimizes the KL divergence between representations and embeddings, effectively encouraging the encoder to internalize the learned structure while preserving the projector's role in loss optimization.

We evaluate SSD across diverse federated learning scenarios, including cross-silo (few clients with large datasets) and cross-device (many clients with small datasets) settings. Our experiments span both in-distribution tasks and out-of-distribution (OOD) datasets to assess generalizability. Results demonstrate that SSD consistently outperforms existing methods in both downstream task performance (measured via linear probing and fine-tuning accuracy) and representation quality (quantified using effective rank~\cite{garrido2023rankme} and uniformity~\cite{wang2020alignuniform} metrics).

Our main contributions are threefold:
\begin{itemize}
    \item We identify and formalize the challenge of inter-client uniformity in FUL, establishing it as a critical direction for decentralized unsupervised representation learning.
    \item  We propose Soft Separation and Distillation (SSD), a simple yet effective framework that addresses inter-client uniformity without additional communication overhead or compromising privacy.
    \item  We conduct extensive experiments across varied FL settings and tasks, confirming SSD's superiority over baseline methods in both performance and robustness.
\end{itemize}

\section{Preliminaries}
\vspace{5pt}
\noindent\textbf{Federated unsupervised learning (FUL)} is a learning paradigm in which multiple clients collaboratively train a model without sharing new raw data, and where the local datasets contain unlabeled data. The goal is to learn a global representation that generalizes across all participating clients. 
Consider a federated learning system with $K$ local  clients and a central server. Each client $k$ has access to an unlabeled local dataset $\mathcal{D}_k$, modeled as samples from a client-specific data distribution $p_k(\mathbf{x})$:
\begin{align}
    \mathcal{D}_k = \{ \mathbf{x}_i^k\}_{i=1}^{n_k} \sim p_k(\mathbf{x}),
\end{align}
where $n_k$ denotes the number of samples in $\mathcal{D}_k$.
The global objective in FUL can be formulated as:
\begin{align}
    \min_{\theta} \sum_{k=1}^K P_k \mathbb{E}_{\mathbf{x} \sim p_k(\mathbf{x})} \big[\mathcal{L}_k(\theta; \mathbf{x)\big]},
\end{align}
where $P_k$ is the probability of drawing a data from client $k$, and $\mathcal{L}_k(\cdot)$ denotes the local unsupervised loss of client $k$.
\vspace{5pt}

\noindent\textbf{Self-Supervised Representation Learning} aims to learn meaningful representations that capture the underlying structure of data. A typical framework consists of the following components: 

\begin{itemize}
    \setlength\itemsep{1em}
    \item \textbf{Encoder} $f(\cdot)$: Maps (augmented) input data $\tilde{\mathbf{x}} \sim \mathcal{T}(\mathbf{x})$ into a \emph{representation} $\mathbf{h} = f(\tilde{\mathbf{x}}) \in \mathbb{R}^d$, where $d$ is the hidden dimension. It is common to use a deep neural network (\eg{}, ResNet50) as the encoder. 
    \item \textbf{Projector} $g(\cdot)$: Transforms the representation $\mathbf{h}$ into an \emph{embedding} $\mathbf{z} = g(\mathbf{h}) \in \mathbb{R}^d$. The training loss function is usually applied on this embedding space, and the projector is often a small network such as a multilayer perceptron.
\end{itemize}

A fundamental principle in self-supervised representation learning is that similar samples should have similar representations. In other words, representations should be invariant to minor variations in input data. This is typically achieved by aligning the representations of two augmented versions of the same image (referred to as a positive pair) using the following alignment loss:
\begin{align}
    \mathcal{L}_{\text{align}} = \mathbb{E}_{\mathbf{x}\sim p(\mathbf{x}), \tilde{\mathbf{x}},\tilde{\mathbf{x}}^+ \sim \mathcal{T}(\mathbf{x})} ||\mathbf{z} - \mathbf{z}^+||^2_2,
    \label{eq:align}
\end{align}
where $\tilde{\mathbf{x}},\tilde{\mathbf{x}}^+ \sim \mathcal{T}(\mathbf{x})$ denotes two independent augmentations of the same sample $\mathbf{x}$ and $\mathbf{z} = g(f(\tilde{\mathbf{x}})), \mathbf{z}^+ = g(f(\tilde{\mathbf{x}}^+))$ are the corresponding embeddings.

Although aligning positive pairs encourages similarity, it can lead to a degenerate \emph{collapse} where all representations converge to a single point. To circumvent this, \citet{wang2020alignuniform} introduces a \emph{uniformity} objective to ensure that features are evenly dispersed on the unit hypersphere. This uniformity is often measured using the log of the average Gaussian potential~\cite{cohn2007universally}:
\begin{align}
    \mathcal{L}_{\text{uniform}} = - \log \mathbb{E}_{\mathbf{z}_i, \mathbf{z}_j \overset{\text{i.i.d}}{\sim} p(\mathbf{z})}\bigl[e^{-t ||\mathbf{z}_i - \mathbf{z}_j||^2_2}\bigr] 
    \label{eq:uniform}
\end{align}

where $p(\mathbf{z})$ is the distribution of embeddings obtained by mapping data samples through the encoder $f$ and projector $g$ and $t$ is the temperature hyperparameter.
\section{Methodology}
\label{sec:method}

In this section, we introduce \emph{Soft Separation and Distillation (SSD)}, a novel approach designed to enhance representation uniformity in federated learning (FL). We first analyze the limitations of uniformity in non-IID FL settings, then propose dimension-scaled regularization to softly separate client features, and finally introduce projector distillation  to effectively transfer improved embedding uniformity to the representations.

\begin{figure*}
    \centering
    \includegraphics[width=1\linewidth, trim={3cm 0 1.5cm 0cm} ]{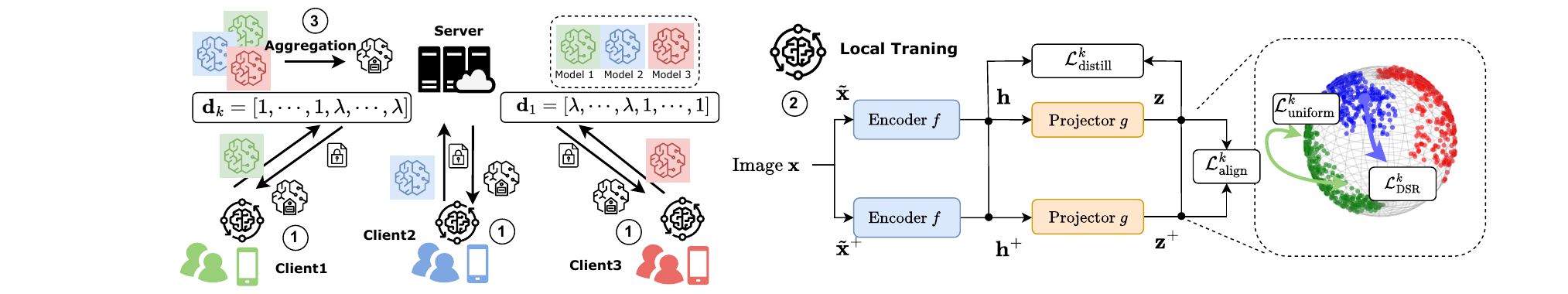}
    \vspace{-10px}
    \caption{\textbf{Overview of our Training Pipeline.} \circled{1} The server \emph{initially} assigns distinct weight vectors to each client. \circled{2} Clients perform local training by optimizing their respective loss functions on their own data. \circled{3} The server aggregates updated parameters from all clients. Steps \circled{2} and \circled{3} are iteratively repeated until convergence.} 
    \label{fig:fl_overview}
    \vspace{-10px}
\end{figure*}

\vspace{0.1cm}
\noindent\textbf{Limited Inter-Client Uniformity of FL.}
\label{sec:limited_uniformity}
Uniformity is a critical metric in representation learning~\cite{wang2020alignuniform, wang2021behavior, fang2024uniformity}, where higher uniformity indicates better preservation of information in the learned representations. In centralized training, jointly optimizing alignment loss $\mathcal{L}_{\text{align}}$ and uniformity loss $\mathcal{L}_{\text{uniform}}$ yields high-quality representations.
However, the distributed nature of FL introduces a fundamental challenge to uniformity optimization. Specifically, the uniformity metric can be decomposed into intra-client and inter-client components:
\begin{equation}
\begin{aligned}
    \mathcal{L}_{\text{uniform}}
    = - \log \biggl( \underbrace{\sum_{k=1}^K \mathbb{E}_{\mathbf{z}, \mathbf{z}' \overset{\text{i.i.d}}\sim p_k(\mathbf{z})} [e^{-t {||\mathbf{z} - \mathbf{z}'||^2_2}}] }_{\text{intra-client }  \mathcal{L}_{\text{uniform}}^k} \\ + \underbrace{\sum_{i\neq j} \mathbb{E}_{\mathbf{z} \sim p_i(\mathbf{z}), \mathbf{z}'\sim p_j(\mathbf{z})}[e^{-t {||\mathbf{z} - \mathbf{z}'||^2_2}}]}_{\text{inter-client }\mathcal{L}_{\text{uniform}}^{(i, j)}}  \biggr) ,
\end{aligned}
\label{eq:uniform_decompose}
\end{equation}
where the first term represents intra-client uniformity $\mathcal{L}_{\text{uniform}}^k$ and the second term represents inter-client uniformity $\mathcal{L}_{\text{uniform}}^{(i,j)}$.

In federated learning, each client $i$ optimizes its local loss function using samples drawn from its own local distribution $p_i$, which means that only the intra-client term in the above equation is explicitly optimized. In a homogeneous setting, when client distributions are similar (i.e., $p_i \approx p_j$), optimizing intra-client uniformity naturally promotes good inter-client uniformity. However, in non-IID settings where client distributions differ significantly, optimizing only intra-client uniformity does not guarantee good inter-client uniformity, potentially limiting the quality of the globally aggregated model. The core challenge in enhancing inter-client uniformity lies in the federated learning constraint that the server has no access to raw client data or embeddings, making it impossible to directly impose a loss function that operates across different clients.

\begin{figure*}
    \centering
    \includegraphics[width=\linewidth]{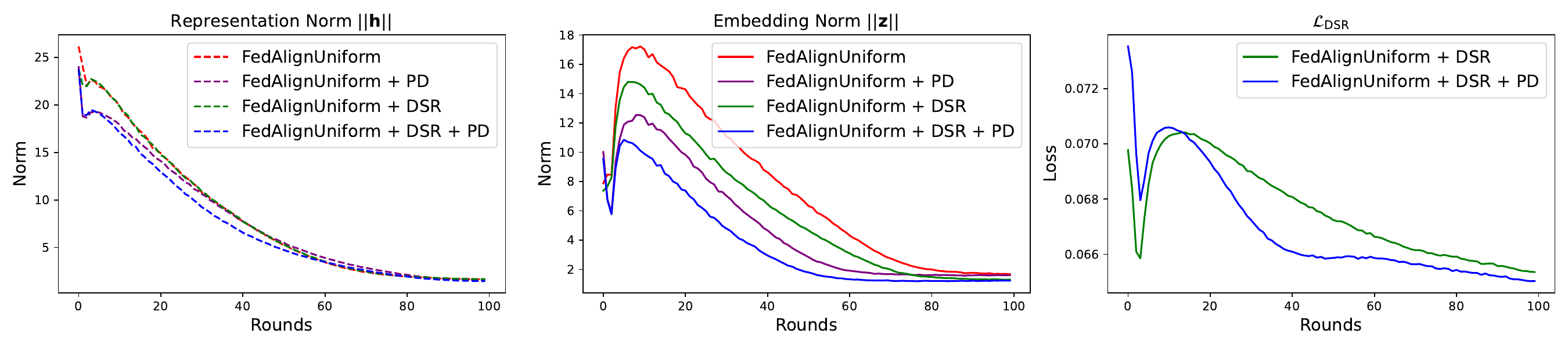}
    \caption{\textbf{Representation norm, embedding norm, and DSR loss during training.} Adding DSR primarily affects the embedding space, as reflected in the embedding norm, while leaving the representation norm unchanged. In contrast, incorporating PD directly influences the representation norm, and when combined with DSR, it leads to a more pronounced decrease in representation norm. Additionally, PD enhances the effect of DSR, resulting in a lower $\mathcal{L}_{\text{DSR}}$.}
    \label{fig:feature_norm}
    \vspace{-10px}
\end{figure*}


\vspace{0.1cm}
\noindent\textbf{Dimension-Scaled Regularization (DSR).}
\label{sec:dsr}
To address the inter-client uniformity challenge, we propose Dimension-Scaled Regularization (DSR), which improves global uniformity by encouraging client embeddings to spread in different directions of the representation space. This approach effectively increases the separation between clients without enforcing rigid boundaries that might distort the intrinsic data structure.

For each client $k$, we define a dimension-scaling vector $\mathbf{d}_k \in \mathbb{R}^d$, where $d$ is the dimensionality of the embedding space. This vector applies selective scaling to specific dimensions, with some dimensions scaled by a factor $\alpha \neq 1$ in a client-specific manner:
\begin{equation}
d_{i,k} = 
\begin{cases}
\alpha, & \text{if } i \in \mathcal{S}_k  \\
1, & \text{{otherwise}},\\
\end{cases}
\end{equation}

where $\mathcal{S}_k$ is a set of dimensions uniquely assigned to client $k$, ensuring that the scaled dimensions are non-overlapping across clients (i.e., $\mathcal{S}_i \cap \mathcal{S}_j = \emptyset$ for $i \neq j$). The size of each set $\mathcal{S}_k$ is approximately $\lfloor d/K \rfloor$, where $K$ is the number of clients.


We then regularize each client's embeddings by encouraging them to move toward their dimension-scaled versions through the following loss:
\begin{equation}
    \mathcal{L}_{\text{DSR}}^k = \mathbb{E}_{\mathbf{z}\sim p_k(\mathbf{z})}\left[\|\mathbf{z} - \text{stopgrad}(\mathbf{z} \odot \mathbf{d}_k)\|_2^2\right],
\end{equation}
where $\odot$ represents element-wise multiplication, and $\text{stopgrad}(\cdot)$ prevents gradient flow through the scaled target, ensuring that we pull the original embedding toward the scaled version rather than vice versa.

\begin{figure}
    \centering
\includegraphics[trim={3.3cm 0 2.3cm 0},clip, width=0.8\linewidth]{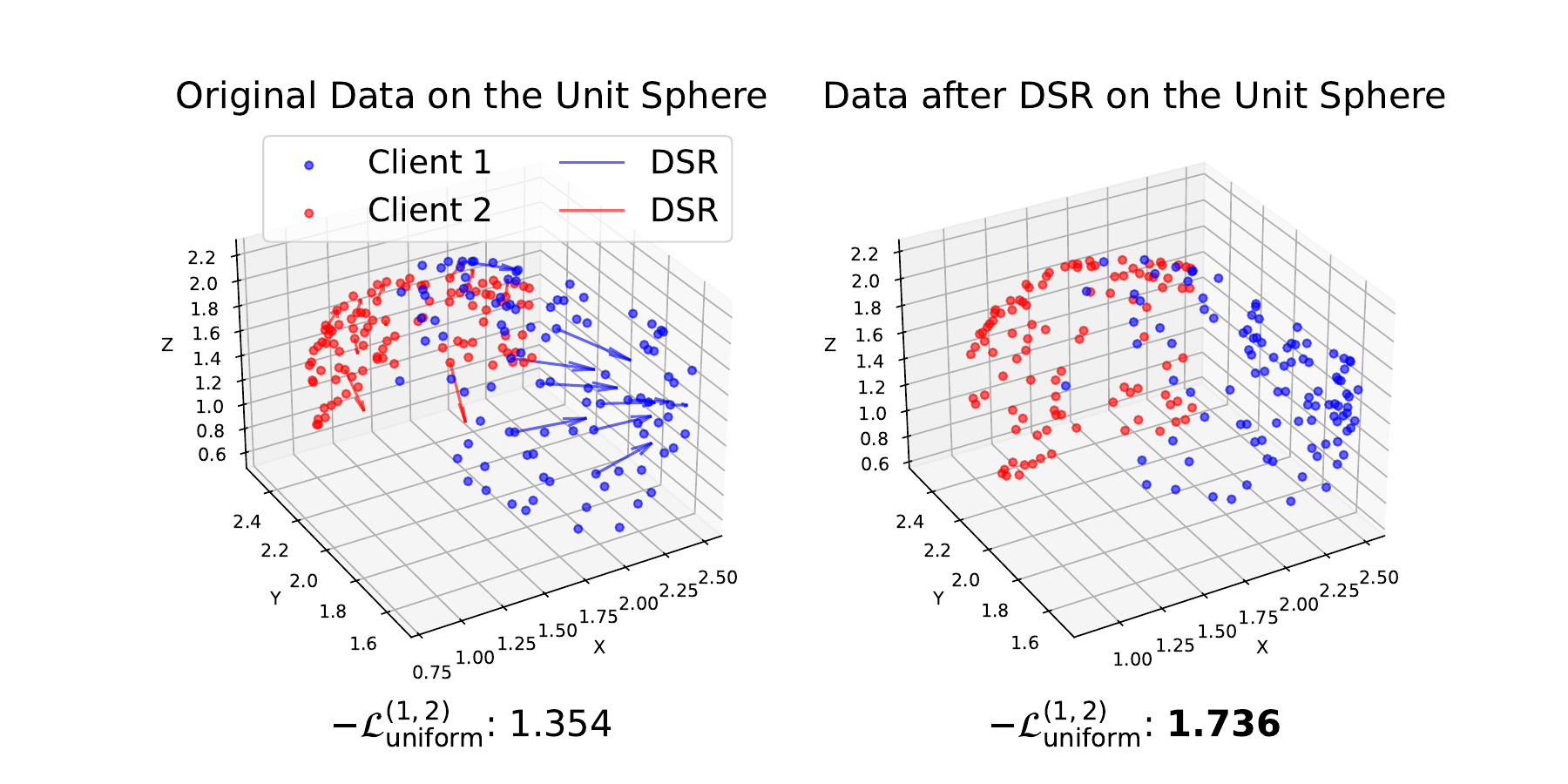}
    \caption{\textbf{Illustration of the effect of DSR.} The left figure shows the original data, where arrows indicate the transformation direction after applying DSR. The right figure presents the data after transformation. DSR increases separation and enhances inter-client uniformity.}
    \label{fig:dsr_demo}
    \vspace{-10px}
\end{figure}

To understand why DSR enhances inter-client uniformity, we can analyze its effect on the representation distribution. When a vector $\mathbf{z}$ is scaled along specific dimensions by a factor $\alpha > 1$ and then normalized to the unit hypersphere, the resulting vector shifts toward those scaled dimensions. By assigning different scaling dimensions to each client, DSR effectively encourages client representations to occupy different regions of the unit hypersphere, as illustrated in Figure~\ref{fig:dsr_demo}. Mathematically, if clients $i$ and $j$ have scaling vectors $\mathbf{d}_i$ and $\mathbf{d}_j$ with non-overlapping scaled dimensions, their optimized embeddings will tend to have a smaller dot product:
\begin{equation}
    \mathbb{E}_{\mathbf{z}_i \sim p_i(\mathbf{z}), \mathbf{z}_j \sim p_j(\mathbf{z})}[\mathbf{z}_i^\top \mathbf{z}_j] < \mathbb{E}_{\mathbf{z}_i, \mathbf{z}_j \sim p(\mathbf{z})}[\mathbf{z}_i^\top \mathbf{z}_j],
\end{equation} where $p(\mathbf{z})$ represents the distribution without DSR. This reduced dot product corresponds to increased angular separation, directly contributing to improved uniformity across the global distribution.

Unlike a hard separation approach that would restrict each client to entirely separate subspaces (which could severely constrain representation capacity), our soft separation allows clients to share most dimensions while gently pushing them toward different directions. This balanced approach preserves the flexibility needed for effective representation learning while significantly improving inter-client uniformity.

An extreme example of fully separating embeddings would be assigning each client a dedicated subspace (e.g., if $d=500$ and $K=10$, each client could occupy 50 distinct dimensions). Although this configuration maximizes inter-client uniformity, it may disrupt the intrinsic data structure and miss out on the collaborative benefits of FL. In contrast, our \emph{soft} DSR loss encourages some degree of separation without completely isolating the clients' embeddings. We analyze this tradeoff further in Section~\ref{sec:analysis}.

\vspace{0.1cm}
\noindent\textbf{Projector Distillation (PD).}
\label{sec:pd}
While DSR effectively enhances uniformity at the embedding level, we observe that this improvement may not fully transfer to the representation level. This discrepancy occurs because the projector $g(\cdot)$ placed between the encoder $f(\cdot)$ and the loss function can absorb much of the optimization effect, as shown in Figure~\ref{fig:feature_norm}.  Although completely removing the projector might seem like a direct solution, prior work~\cite{chen2020simclr, gupta2022projectionhead, xue2024projectionhead} has demonstrated that the projector plays a crucial role in separating optimization objectives from feature representations, thereby preventing overfitting to specific self-supervised tasks and improving downstream performance. 
 
 To bridge this gap, we introduce Projector Distillation (PD), which explicitly aligns the encoder's representations with the projector's embeddings:
\begin{equation}
    \mathcal{L}_{\text{distill}}^k = \mathbb{E}_{x\sim p_k(\mathbf{x})}\left[D_{\text{KL}}\left(\sigma(\mathbf{h}) \| \sigma(\mathbf{z})\right)\right],
\end{equation}
where $\mathbf{h} = f(\mathbf{x})$ is the representation from the encoder, $\mathbf{z} = g(\mathbf{h})$ is the embedding from the projector, $\sigma(\cdot)$ denotes the softmax function, and $D_{\text{KL}}$ is the Kullback-Leibler divergence.

This distillation mechanism encourages the encoder to internalize the beneficial structure learned in the embedding space, effectively transferring the improved uniformity from embeddings to representations. By minimizing the KL divergence between these distributions, we ensure that the uniformity improvements achieved through DSR are reflected in the representations used for downstream tasks.

\SetKwInput{KwInput}{Input}                
\SetKwInput{KwOutput}{Output}              
\SetKwFunction{FnClient}{}
\SetKwFunction{FnServer}{}

\begin{algorithm}[t]
\caption{\small {\emph{Soft Separation and Distillation (\Method)}}}
    \label{algo:ssd}
    \SetAlgoLined
    \KwInput{communication rounds $T$, local epochs $E$, number of clients $K$, dataset $\mathcal{D}=\cup_{k\in[K]} \mathcal{D}_k$, augmentation function $\mathcal{T}(\cdot)$}
    \KwOutput{Final global encoder $f_\theta^T$}

    \SetKwProg{Fn}{Server}{:}{}
    \Fn{\FnServer}{
        Initialize global encoder $f_\theta^0$ and projector $g_\theta^0$\;
        Assign weight vector \textcolor{red}{$\{\mathbf{d}_k\}_{k \in [K]}$} for all clients\;
        \For{$t = 0$ \textbf{to} $T-1$}{
            $S_t \leftarrow$ (randomly select a set of clients)\;
            \For{each client $k \in S_t$ in parallel}{
                $f_\theta^{(t,k)}, g_\theta^{(t,k)} \leftarrow$ \textbf{Client}($f_\theta^t$, $g_\theta^t$, $\mathbf{d}_k$)\;
            }
            \textcolor{BlueViolet}{// Model aggregation}\;
            \For{each parameter $\theta$ in $f, g$}{
                $\theta^{t+1} \leftarrow \sum_{k \in S_t} w_k \theta^{(t,k)}$ 
            }
        }
        \KwRet $f_\theta^{T}$\;
    }

    \SetKwProg{Fn}{Client}{:}{\KwRet}
    \Fn{\FnClient{$f_\theta^t$, $g_\theta^t$, $\mathbf{d}_k$}}{
        \textcolor{BlueViolet}{// Initialize local models}\;
        $f \leftarrow f_\theta^t$, $g \leftarrow g_\theta^t$\;
        
        \For{\textbf{each local epoch} $e = 0$ \textbf{to} $E-1$}{
            \For{\textbf{each batch} $\{\mathbf{x}_i\}_{i=1}^{n_B} \in \mathcal{D}_k$}{
                \textcolor{BlueViolet}{// Data augmentation}\;
                $\tilde{\mathbf{x}}_i, \tilde{\mathbf{x}}_i^+ = \mathcal{T}(\mathbf{x}_i), \mathcal{T}(\mathbf{x}_i)$ \;

                \textcolor{BlueViolet}{// Get representations and embeddings}\;
                $\mathbf{h}_i, \mathbf{h}_i^+ = f(\tilde{\mathbf{x}}_i), f(\tilde{\mathbf{x}}_i^+)$ \;
                $\mathbf{z}_i, \mathbf{z}_i^+ = g(\mathbf{h}_i), g(\mathbf{h}_i^+)$ \;


                \textcolor{BlueViolet}{// Optimize model using combined loss}\;
                Update $f, g$ by minimizing \textcolor{red}{$\mathcal{L}^k$} (Eq~\eqref{eq:total})\;
            }
        }
        \KwRet $f, g$ \textcolor{BlueViolet}{// Model upload}\; 
    }
\end{algorithm}

\subsection{Overview of Training Pipeline}
\label{sec:algo_overview8}
The complete SSD algorithm is outlined in Algorithm~\ref{algo:ssd}, with our key design components highlighted in red. The server initially assigns unique weight vectors to all clients for dimension-scaled regularization. During local training, each client optimizes a combination of the standard alignment and uniformity losses for self-supervised learning, augmented with our DSR and PD terms. The local training objective for each client $k$ is formulated as:

\begin{equation}
    \mathcal{L}^k = \mathcal{L}_{\text{align}}^k + \beta\mathcal{L}_{\text{uniform}}^k + \gamma\mathcal{L}_{\text{DSR}}^k + \delta\mathcal{L}_{\text{distill}}^k,
    \label{eq:total}
\end{equation}
where $\beta$, $\gamma$, and $\delta$ are hyperparameters that balance the contribution of each loss term.

After local training, clients upload their updated models to the server, which aggregates them using standard federated averaging~\cite{mcmahan2017fedavg}: $\theta^{t+1} = \sum_{k \in S_t} w_k \theta^{(t,k)}$ where $w_k = \frac{n_k}{\sum_{i=1}^K n_i}$ represents the aggregation weight based on the client's data size.

Importantly, our method preserves privacy since it requires no sharing of raw data. The only additional communication is the initial weight vectors $\{\mathbf{d}_k\}_{k=1}^K$, which are lightweight and contain no sensitive information. The computational overhead of SSD is minimal, making it highly practical for real-world federated learning deployments.
\section{Experiments}

\begin{table*}[t]
\caption{\textbf{Results on Cross-Silo and Cross-Device Settings.} Accuracy (\%) of linear probing (LP), fine-tuning (FT) 1\%, and 10\% labeled data on CIFAR10 and CIFAR100 ($\boldsymbol{\alpha}=0.1$). The highest score is highlighted in \textbf{bold}, and the second-highest score is \underline{underlined}.}
\resizebox{\linewidth}{!}{
\begin{tabular}{ccccccccccccc}
\toprule
 & \multicolumn{6}{c|}{\bf CIFAR10} & \multicolumn{6}{c}{\bf CIFAR100} 
\\
 & \multicolumn{3}{c|}{Cross-Silo (K=10)} & \multicolumn{3}{c|}{Cross-Device (K=50)} & \multicolumn{3}{c|}{Cross-Silo (K=10)} & \multicolumn{3}{c}{Cross-Device (K=50)} \\
 & LP & FT 1\% & \multicolumn{1}{c|}{FT 10\%} & LP & FT 1\% & \multicolumn{1}{c|}{FT 10\%} & LP & FT 1\% & \multicolumn{1}{c|}{FT 10\%} & LP & FT 1\% & FT 10\% \\
\midrule
FedAlignUniform~\cite{wang2020alignuniform} & 80.84 & \underline
{69.99} & 81.00 & 71.28 & 57.41 & 73.77 & 57.25 & 28.97 &	48.99	& 43.03 &	16.37& 36.64\\ 
FedX~\cite{han2022fedx} & 78.4 & 66.78 & 80.01 &  71.01 & 56.91 & 73.24 & 57.34 & 27.46 & 49.50 & 43.07 & 16.04 & 35.46\\ 
L-DAWA~\cite{rehman2023ldawa} & 77.67& 65.94 &	79.34	&					67.65	& 53.75	& 71.22 & 56.90 & 27.08 & 	49.57 & 42.58& 15.54 & 	34.82 \\
FedDecorr~\cite{shi2023feddecorr} & 80.13 & 69.09& 80.33 & \underline
{71.49} & \bf  58.19& 73.97 & 57.25	& 29.38	&49.53	& \underline
{44.74} & \underline
{17.67} & \underline{36.68}  \\
FedU2~\cite{liao2024fedu2} & \underline
{81.01} & 69.62& \underline
{81.01} & 71.09	& 57.15& \underline
{74.21} &  \bf 57.40	&\underline
{29.39}	& \underline
{49.64}	& 42.90 & 16.08	& 35.48\\
\bf \Method & \bf 81.32 & \bf 	70.74& \bf 
 81.67 & \bf  71.83& \underline
{57.77} & \bf 74.61 & 
 \underline
{57.38} & \bf 29.57 & \bf 49.87 & \bf 45.21 &	\bf 17.70 & \bf 36.82\\
\bottomrule
\end{tabular}
}
\label{tab:fl}
\end{table*}

\noindent\textbf{Datasets.} We evaluate our approach on CIFAR-10 and CIFAR-100, which contain 50K training images and 10K testing images, with 10 and 100 classes, respectively. Following prior work~\cite{zhuang2022fedema, liao2024fedu2}, we simulate non-IID data across $K$ clients using a Dirichlet prior $\text{Dir}(\boldsymbol{\alpha})$, where a smaller $\boldsymbol{\alpha}$ indicates a higher degree of data heterogeneity. We conduct experiments in two federated learning settings: cross-silo ($K=10$) with full participation and cross-device ($K=50$) with a participation rate of 0.2.
\vspace{3pt}

\noindent\textbf{Evaluation Metrics.} We evaluate our approach based on downstream performance and representation quality. For downstream performance, we follow prior work~\cite{zhuang2021fedu, han2022fedx, liao2024fedu2}, using linear probing and fine-tuning with 1\% and 10\% of the data. For representation quality, we assess uniformity~\cite{wang2020alignuniform, wang2021behavior, fang2024uniformity} as defined in Equation~\eqref{eq:uniform} and effective rank~\cite{roy2007effrank, garrido2023rankme, zhang2023gcf} (see the definition in Appendix~\ref{app:eff_rank}).


\vspace{3pt}

\noindent\textbf{Baselines.} 
We compare our method against three categories of baselines. (1) Adapting a centralized algorithm to the federated setting: FedAlignUniform~\cite{wang2020alignuniform}. (2) FUL methods that mitigate divergence from the global model: FedX~\cite{han2022fedx} and L-DAWA~\cite{rehman2023ldawa}. (3) FUL methods addressing dimensional collapse in local clients: FedDecorr~\cite{shi2023feddecorr}. FedU2~\cite{liao2024fedu2} incorporates modules that tackle both divergence control and dimensional collapse.

\vspace{3pt}

\noindent\textbf{Implementation Details.} We follow prior work~\cite{liao2024fedu2} for image augmentations, training configurations, and model architecture, using ResNet-18~\cite{he2016resnet} as the encoder and a two-layer linear projector. The hyperparameters $\beta$, $\gamma$, and $\delta$ are set to 1.0, 1.0, and 0.1, respectively, and the scaling factor $\alpha$ is set to 10.0. Further details are provided in Appendix~\ref{app:training}.

\subsection{Main Results} 

\noindent\textbf{Performance on different FL settings.} Table~\ref{tab:fl} compares the performance of various methods in cross-silo and cross-device federated learning. SSD consistently achieves the highest accuracy and excels in fine-tuning tasks. FedAlignUniform, FedDecorr, and FedU2 are competitive but generally lag behind SSD, while FedX and L-DAWA—aimed at ensuring global model consistency—perform the worst across all settings. Although FedDecorr and FedU2 enhance intra-client uniformity (with FedU2 adding a consistency module), they still trail SSD. By improving inter-client uniformity, SSD underscores the importance of addressing inter-client variations in federated learning.

\vspace{3pt}
\noindent\textbf{Representation quality.} Figure~\ref{fig:effrank_unifority} compares different methods based on effective rank~\cite{roy2007effrank} and uniformity~\cite{wang2020alignuniform}, two key metrics for representation quality.
Among state-of-the-art methods, FedX and L-DAWA, which emphasize global model consistency, achieve strong representation quality. In contrast, methods focusing on local uniformity, such as FedDecorr and FedU2, exhibit lower effective rank and uniformity, indicating that they fail to address global uniformity.
Our method, SSD, surpasses all other federated approaches, achieving the highest effective rank and the best uniformity. This demonstrates that SSD effectively enhances representation quality, outperforming existing methods.

\vspace{3pt}
\noindent\textbf{Generalization on OOD datasets.}
Evaluating generalization to out-of-distribution (OOD) datasets helps assess whether models can learn transferable representations that perform well beyond their training distribution. Table~\ref{tab:ood} presents the generalization performance of different methods, where models trained on CIFAR-100 and TinyImageNet-200 are evaluated on CIFAR-10.

Among state-of-the-art methods, FedX, which exhibits high representation quality but performs poorly in in-distribution settings, achieves the best generalization in OOD scenarios. However, our method, SSD, consistently outperforms all baselines, achieving the highest accuracy in both settings. SSD also demonstrates the best uniformity~\cite{wang2020alignuniform} and effective rank~\cite{garrido2023rankme}, indicating superior feature representation quality. While FedAlignUniform, FedDecorr, L-DAWA and FedU2 perform competitively, they generally fall short of SSD, further reinforcing its effectiveness in handling distribution shifts.

\subsection{Analysis}
\label{sec:analysis}

\begin{table}
  \centering
  \caption{\textbf{Generalization on OOD datasets.} Accuracy (\%) of linear probing, uniformity, and effective rank when trained on CIFAR-100 and TinyImageNet-200, and evaluated on CIFAR-10.}
  \resizebox{\linewidth}{!}{
  \begin{tabular}{cccc|ccc}
    \toprule
     &  \multicolumn{3}{c|}{\bf CIRAR100 $\rightarrow$ CIFAR10} &  \multicolumn{3}{c}{\bf TinyImageNet $\rightarrow$ CIFAR10} \\
     & LP & $-\mathcal{L}_{\text{uniform}} (\uparrow)$ & ERank ($\uparrow$) & LP & $-\mathcal{L}_{\text{uniform}} (\uparrow)$ & ERank ($\uparrow$)
     \\
    \midrule
    FedAlignUniform~\cite{wang2020alignuniform} & 77.66 & 3.65 & 66.99 & 79.86 & 3.71  & 76.31 \\
    FedX~\cite{han2022fedx} & \underline{78.02} & \bf 3.73 & \underline{84.88} & \underline{79.87} & \underline{3.76} & \underline{93.48} \\ 
L-DAWA~\cite{rehman2023ldawa} & 77.46 & 3.71	& 84.75	& 79.62	& 3.74	& 93.41\\
    FedDecorr~\cite{shi2023feddecorr} & 77.62 & 3.66 & 74.20 & 79.79 & 3.72 & 84.93\\
    FedU2~\cite{liao2024fedu2} & 77.74 & 3.66 & 68.3 & 79.74 & 3.69 & 75.97 \\
    \bf  \Method & \bf 78.48 & \bf 3.73 & \bf 86.95 & \bf 80.00 & \bf 3.77 & \bf 98.45 \\
    \bottomrule
  \end{tabular}
  }
  
  \label{tab:ood}
\end{table}
\noindent\textbf{Ablation Study.} We provide the ablation study in Table~\ref{tab:ablation}. It  shows that adding projector distillation (PD) alone does not provide noticeable improvements. In contrast, adding dimension-scaled regularization (DSR) alone leads to slight improvements in LP and FT 10\%, but the uniformity enhancement is minimal. However, when combining both DSR and PD (SSD), the method effectively transfers optimization benefits to representation quality, leading to a significant uniformity improvement and achieving the best overall performance.

\begin{table}
  \centering
    \caption{\textbf{Ablation Study.} PD alone has minimal impact on performance. Adding DSR alone provides a slight improvement in both performance and uniformity. Combining both DSR and PD leads to a significant boost in uniformity and achieves the best performance.}
  \resizebox{\linewidth}{!}{
  \begin{tabular}{ccccc}
    \toprule
     & LP & FT 1\% & FT 10\%& $-\mathcal{L}_{\text{uniform}} (\uparrow)$  \\
    \midrule
FedAlignUniform~\cite{wang2020alignuniform}  & 80.84 & 69.99 & 81.00 & 3.79\\ 
\quad +PD & 80.74 & 69.78 & 80.71 & 3.80 \\
\quad +DSR & 81.05 & 69.77 & 81.15 & 3.81 \\
\quad + DSR + PD (\textbf{SSD}) & \bf 81.32 & \bf 70.74 & \bf  81.67 & \bf 3.84 \\
    \bottomrule
  \end{tabular}
  }

  \label{tab:ablation}
\end{table}

\begin{figure}[t] 
    \centering
    \begin{subfigure}[b]{0.48\linewidth}
        \centering
        \includegraphics[width=\textwidth, trim={0.2cm 0 0.9cm 0},clip]{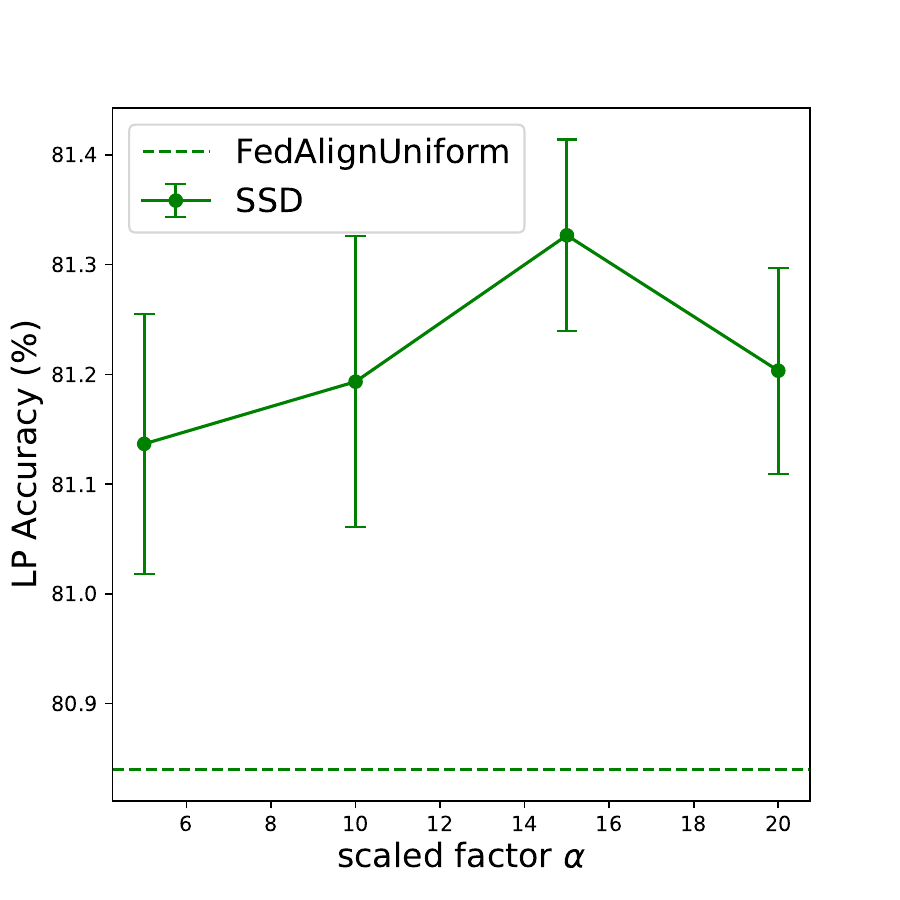} 
        \caption{Robustness}
        \label{fig:robustness}
    \end{subfigure}
    \hfill
    \begin{subfigure}[b]{0.48\linewidth}
        \centering
        \includegraphics[width=\textwidth, trim={0.2cm 0 0.9cm 0},clip]{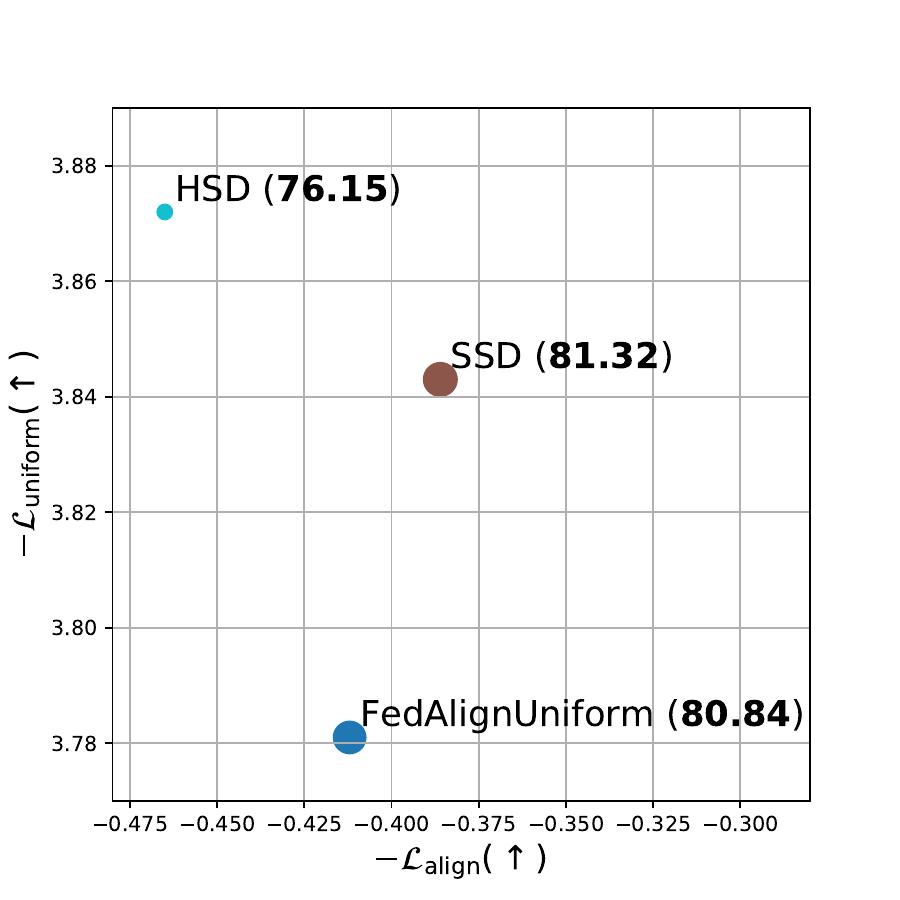} 
        \caption{Separation}
        \label{fig:separation}
    \end{subfigure}
    
    \caption{(a) \textbf{Robustness of DSR} across different scaled factors $\alpha$, where each $\alpha$ is tested with three different randomly selected scaled dimensions. (b) \textbf{Soft \vs{} hard separation.} HSD achieves the highest uniformity but at the cost of reduced alignment, resulting in poor overall performance.}
    \label{fig:main}
\end{figure}

\vspace{3pt}
\noindent\textbf{Robustness of DSR.} We evaluate our method from two perspectives: the scaled factor $\alpha$ and the selection of scaled dimensions for each client. The scaled factor $\alpha$ influences the overall discrepancy, while the selection of scaled dimensions affects the direction each client is guided toward. To assess robustness, each $\alpha$ is evaluated with three different selections of scaled dimensions. As shown in Figure~\ref{fig:robustness}, our method maintains stable performance across different $\alpha$ values, introducing minimal variation while consistently outperforming FedAlignUniform. This demonstrates that our approach effectively balances alignment and uniformity without being overly sensitive to the choice of $\alpha$ or the specific dimension selection.

\vspace{3pt}
\noindent\textbf{Soft \vs{} hard client separation.} We examine the impact of client feature separation on alignment, uniformity, and overall performance. Our intuition is that increasing separation between clients enhances global uniformity. One way to enforce this separation is by restricting each client to its own subspace without sharing information with others. To illustrate the effects of such strict partitioning, we consider Hard Separation and Distillation (HSD) as a baseline. As shown in Figure~\ref{fig:separation}, while HSD achieves the highest uniformity, it does so at the cost of severely reduced alignment, ultimately leading to poor downstream performance. This performance drop can be attributed to the disruption of intrinsic feature structures. In contrast, our proposed SSD balances both alignment and uniformity, mitigating the drawbacks of hard separation while enhancing uniformity, resulting in improved overall performance.

\vspace{3pt}
\noindent\textbf{Why not remove projector or apply loss on representations?}
A natural idea for addressing the limited transfer of embedding-level loss optimization (in terms of representation uniformity) is to remove the projector or apply the loss directly to the representations. However, prior work \cite{chen2020simclr, gupta2022projectionhead, xue2024projectionhead} has demonstrated that the projector plays a critical role in preventing the encoder from overfitting on the upstream task. Indeed, as shown in Table~\ref{tab:projector}, removing the projector and adding DSR does substantially improve uniformity and downstream performance, but these results still lag behind those achieved when the projector is retained. Therefore, it is crucial to develop a method that effectively promotes uniformity while preserving the benefits offered by the projector.

\begin{table}
  \centering
    \caption{\textbf{The Effect of the Projector.} Without the projector, DSR significantly improves uniformity and performance. Yet, the overall results remain inferior to those achieved with the projector.}
  \resizebox{0.8\linewidth}{!}{
  \begin{tabular}{cccc}
    \toprule
     & Projector & LP & $-\mathcal{L}_{\text{uniform}} (\uparrow)$  \\
    \midrule
FedAlignUniform  & \xmark & 73.16  & 3.72 \\ 
\quad + DSR  & \xmark & 76.14 \textcolor{teal}{\small (+2.98)}  & 3.77 \textcolor{teal}{\small (+0.05)} \\ 
\midrule
FedAlignUniform  & \cmark & 80.84  & 3.79 \\ 
\quad + DSR  & \cmark & 81.05 \textcolor{teal}{\small (+0.21)}  & 3.81 \textcolor{teal}{\small (+0.02)}  \\ 
    \bottomrule
  \end{tabular}
  }

  \label{tab:projector}
\end{table}



\section{Related Work}

\subsection{Federated Learning with Non-IID Data}
Federated learning (FL) enables collaborative model training across multiple clients without centralizing data. Since client data distributions are typically non-IID, model convergence and performance are challenging. FedAvg~\cite{mcmahan2017fedavg}, which aggregates local model updates using simple averaging, is the foundational framework but suffers under heterogeneous data and imbalanced client participation.

To address these issues, various aggregation improvements have been proposed. FedProx~\cite{li2020fedprox} reduces client drift by adding a proximal term to limit local updates from straying from the global model. SCAFFOLD~\cite{karimireddy2020scaffold} corrects client drift using control variates to reduce variance in local updates. MOON~\cite{li2021moon} improves representation consistency with a contrastive loss. Other methods use globally shared data to improve generalization; FedShare~\cite{zhao2018fedshare} aligns client models using shared data, while FedDistill~\cite{jeong2018communication} distills knowledge from the global model to local models using shared data.

\subsection{Federated Unsupervised Learning}
Federated Unsupervised Learning (FUL) combines federated learning with un-/self-supervised learning, primarily addressing (1) non-IID data distribution and (2) representation collapse.

To handle non-IID data, FedCA~\cite{zhang2023fedca} uses a shared dictionary module for better aggregation but risks privacy leakage. FedU~\cite{zhuang2021fedu} reduces this risk by selectively uploading the online network’s encoder and deciding predictor updates based on divergence. FedEMA~\cite{zhuang2022fedema} extends this with exponential moving average updates. FedX~\cite{han2022fedx} adds an alignment/contrastive term with the global model, while Orchestra~\cite{lubana2022orchestra} preserves global structural consistency using global centroids. FedU2~\cite{liao2024fedu2} ensures balanced updates across clients for better alignment but does not guarantee global representation uniformity under non-IID settings.

For representation collapse, FedDecorr~\cite{shi2023feddecorr} shows that local clients suffer from dimensional collapse, which propagates to the global model, and addresses it with a local decorrelation loss. FedU2~\cite{liao2024fedu2} encourages local representation uniformity by minimizing divergence with a spherical Gaussian. Our work extends this by showing that intra-client uniformity alone is insufficient inter-client uniformity must also be explicitly addressed.

\section{Conclusion}

We introduce Soft Separation \& Distillation (SSD), a framework designed to enhance representation quality by improving inter-client uniformity in federated learning. SSD consists of a dimension-scaled regularization term that softly separates client embeddings while preserving the intrinsic data structure, and a projector distillation term that transfers the optimization benefits of the projector to the encoder, thereby improving representation quality. SSD achieves state-of-the-art performance in both representation learning and downstream tasks across various training and FL settings. Our work highlights the importance of global representation quality in federated unsupervised learning, opening new directions for future research. 

\section*{Acknowledgment}
We thank the anonymous reviewers and members of CLLab for their constructive feedback. This work is partially supported by the National Science and Technology Council in Taiwan via NSTC 113-2634-F-002-008, 114-2221-E-002-102-MY3, and the National Taiwan University Center for Data Intelligence via NTU-114L900901 and Research Grants Council of the Hong Kong Special Administrative Region, China (CUHK 2410072, RGC R1015-23). We thank to National Center for High-performance Computing (NCHC) of National Applied Research Laboratories (NARLabs) in Taiwan for providing computational and storage resources. H.-T. Lin is honored to be supported by the Leap Fellowship of the Foundation for the Advancement of Outstanding Scholarship in Taiwan since 2025.

{
    \small
    \bibliographystyle{ieeenat_fullname}
    \bibliography{main}

\begin{thebibliography}{37}
\providecommand{\natexlab}[1]{#1}
\providecommand{\url}[1]{\texttt{#1}}
\expandafter\ifx\csname urlstyle\endcsname\relax
  \providecommand{\doi}[1]{doi: #1}\else
  \providecommand{\doi}{doi: \begingroup \urlstyle{rm}\Url}\fi

\bibitem[Baevski et~al.(2020)Baevski, Zhou, Mohamed, and Auli]{baevski2020wav2vec}
Alexei Baevski, Yuhao Zhou, Abdelrahman Mohamed, and Michael Auli.
\newblock wav2vec 2.0: A framework for self-supervised learning of speech representations.
\newblock \emph{Advances in neural information processing systems}, 33:\penalty0 12449--12460, 2020.

\bibitem[Chen et~al.(2020)Chen, Kornblith, Norouzi, and Hinton]{chen2020simclr}
Ting Chen, Simon Kornblith, Mohammad Norouzi, and Geoffrey Hinton.
\newblock A simple framework for contrastive learning of visual representations.
\newblock In \emph{International conference on machine learning}, pages 1597--1607. PmLR, 2020.

\bibitem[Chen and He(2021)]{chen2021simsiam}
Xinlei Chen and Kaiming He.
\newblock Exploring simple siamese representation learning.
\newblock In \emph{Proceedings of the IEEE/CVF conference on computer vision and pattern recognition}, pages 15750--15758, 2021.

\bibitem[Cohn and Kumar(2007)]{cohn2007universally}
Henry Cohn and Abhinav Kumar.
\newblock Universally optimal distribution of points on spheres.
\newblock \emph{Journal of the American Mathematical Society}, 20\penalty0 (1):\penalty0 99--148, 2007.

\bibitem[Deng et~al.(2009)Deng, Dong, Socher, Li, Li, and Fei-Fei]{deng2009imagenet}
Jia Deng, Wei Dong, Richard Socher, Li-Jia Li, Kai Li, and Li Fei-Fei.
\newblock Imagenet: A large-scale hierarchical image database.
\newblock In \emph{2009 IEEE conference on computer vision and pattern recognition}, pages 248--255. Ieee, 2009.

\bibitem[Fang et~al.(2024)Fang, Li, Sun, and Wang]{fang2024uniformity}
Xianghong Fang, Jian Li, Qiang Sun, and Benyou Wang.
\newblock Rethinking the uniformity metric in self-supervised learning.
\newblock In \emph{The Twelfth International Conference on Learning Representations}, 2024.

\bibitem[Garrido et~al.(2023)Garrido, Balestriero, Najman, and Lecun]{garrido2023rankme}
Quentin Garrido, Randall Balestriero, Laurent Najman, and Yann Lecun.
\newblock Rankme: Assessing the downstream performance of pretrained self-supervised representations by their rank.
\newblock In \emph{International conference on machine learning}, pages 10929--10974. PMLR, 2023.

\bibitem[Grill et~al.(2020)Grill, Strub, Altch{\'e}, Tallec, Richemond, Buchatskaya, Doersch, Avila~Pires, Guo, Gheshlaghi~Azar, et~al.]{grill2020byol}
Jean-Bastien Grill, Florian Strub, Florent Altch{\'e}, Corentin Tallec, Pierre Richemond, Elena Buchatskaya, Carl Doersch, Bernardo Avila~Pires, Zhaohan Guo, Mohammad Gheshlaghi~Azar, et~al.
\newblock Bootstrap your own latent-a new approach to self-supervised learning.
\newblock \emph{Advances in neural information processing systems}, 33:\penalty0 21271--21284, 2020.

\bibitem[Gupta et~al.(2022)Gupta, Ajanthan, Hengel, and Gould]{gupta2022projectionhead}
Kartik Gupta, Thalaiyasingam Ajanthan, Anton van~den Hengel, and Stephen Gould.
\newblock Understanding and improving the role of projection head in self-supervised learning.
\newblock \emph{arXiv preprint arXiv:2212.11491}, 2022.

\bibitem[Han et~al.(2022)Han, Park, Wu, Kim, Wu, Xie, and Cha]{han2022fedx}
Sungwon Han, Sungwon Park, Fangzhao Wu, Sundong Kim, Chuhan Wu, Xing Xie, and Meeyoung Cha.
\newblock Fedx: Unsupervised federated learning with cross knowledge distillation.
\newblock In \emph{European Conference on Computer Vision}, pages 691--707. Springer, 2022.

\bibitem[He et~al.(2016)He, Zhang, Ren, and Sun]{he2016resnet}
Kaiming He, Xiangyu Zhang, Shaoqing Ren, and Jian Sun.
\newblock Deep residual learning for image recognition.
\newblock In \emph{Proceedings of the IEEE conference on computer vision and pattern recognition}, pages 770--778, 2016.

\bibitem[He et~al.(2022)He, Chen, Xie, Li, Doll{\'a}r, and Girshick]{he2022mae}
Kaiming He, Xinlei Chen, Saining Xie, Yanghao Li, Piotr Doll{\'a}r, and Ross Girshick.
\newblock Masked autoencoders are scalable vision learners.
\newblock In \emph{Proceedings of the IEEE/CVF conference on computer vision and pattern recognition}, pages 16000--16009, 2022.

\bibitem[Jeong et~al.(2018)Jeong, Oh, Kim, Park, Bennis, and Kim]{jeong2018communication}
Eunjeong Jeong, Seungeun Oh, Hyesung Kim, Jihong Park, Mehdi Bennis, and Seong-Lyun Kim.
\newblock Communication-efficient on-device machine learning: Federated distillation and augmentation under non-iid private data.
\newblock \emph{arXiv preprint arXiv:1811.11479}, 2018.

\bibitem[Jing et~al.(2022)Jing, Vincent, LeCun, and Tian]{jing2022dc}
Li Jing, Pascal Vincent, Yann LeCun, and Yuandong Tian.
\newblock Understanding dimensional collapse in contrastive self-supervised learning.
\newblock In \emph{ICLR}, 2022.

\bibitem[Kairouz et~al.(2021)Kairouz, McMahan, Avent, Bellet, Bennis, Bhagoji, Bonawitz, Charles, Cormode, Cummings, et~al.]{kairouz2021fedadvances}
Peter Kairouz, H~Brendan McMahan, Brendan Avent, Aur{\'e}lien Bellet, Mehdi Bennis, Arjun~Nitin Bhagoji, Kallista Bonawitz, Zachary Charles, Graham Cormode, Rachel Cummings, et~al.
\newblock Advances and open problems in federated learning.
\newblock \emph{Foundations and trends{\textregistered} in machine learning}, 14\penalty0 (1--2):\penalty0 1--210, 2021.

\bibitem[Karimireddy et~al.(2020)Karimireddy, Kale, Mohri, Reddi, Stich, and Suresh]{karimireddy2020scaffold}
Sai~Praneeth Karimireddy, Satyen Kale, Mehryar Mohri, Sashank Reddi, Sebastian Stich, and Ananda~Theertha Suresh.
\newblock Scaffold: Stochastic controlled averaging for federated learning.
\newblock In \emph{International conference on machine learning}, pages 5132--5143. PMLR, 2020.

\bibitem[Kim et~al.(2021)Kim, Oh, Kim, Cho, and Yun]{kim2021klmse}
Taehyeon Kim, Jaehoon Oh, Nak~Yil Kim, Sangwook Cho, and Se-Young Yun.
\newblock Comparing kullback-leibler divergence and mean squared error loss in knowledge distillation.
\newblock In \emph{Proceedings of the Thirtieth International Joint Conference on Artificial Intelligence, {IJCAI-21}}, pages 2628--2635. International Joint Conferences on Artificial Intelligence Organization, 2021.
\newblock Main Track.

\bibitem[Levine et~al.(2016)Levine, Finn, Darrell, and Abbeel]{levine2016e2epolicy}
Sergey Levine, Chelsea Finn, Trevor Darrell, and Pieter Abbeel.
\newblock End-to-end training of deep visuomotor policies.
\newblock \emph{Journal of Machine Learning Research}, 17\penalty0 (39):\penalty0 1--40, 2016.

\bibitem[Li et~al.(2021)Li, He, and Song]{li2021moon}
Qinbin Li, Bingsheng He, and Dawn Song.
\newblock Model-contrastive federated learning.
\newblock In \emph{Proceedings of the IEEE/CVF conference on computer vision and pattern recognition}, pages 10713--10722, 2021.

\bibitem[Li et~al.(2020{\natexlab{a}})Li, Sahu, Talwalkar, and Smith]{li2020flchallenge}
Tian Li, Anit~Kumar Sahu, Ameet Talwalkar, and Virginia Smith.
\newblock Federated learning: Challenges, methods, and future directions.
\newblock \emph{IEEE signal processing magazine}, 37\penalty0 (3):\penalty0 50--60, 2020{\natexlab{a}}.

\bibitem[Li et~al.(2020{\natexlab{b}})Li, Sahu, Zaheer, Sanjabi, Talwalkar, and Smith]{li2020fedprox}
Tian Li, Anit~Kumar Sahu, Manzil Zaheer, Maziar Sanjabi, Ameet Talwalkar, and Virginia Smith.
\newblock Federated optimization in heterogeneous networks.
\newblock \emph{Proceedings of Machine learning and systems}, 2:\penalty0 429--450, 2020{\natexlab{b}}.

\bibitem[Liao et~al.(2024)Liao, Liu, Chen, Zhou, Yu, Zhu, Yao, Wang, Zheng, and Tan]{liao2024fedu2}
Xinting Liao, Weiming Liu, Chaochao Chen, Pengyang Zhou, Fengyuan Yu, Huabin Zhu, Binhui Yao, Tao Wang, Xiaolin Zheng, and Yanchao Tan.
\newblock Rethinking the representation in federated unsupervised learning with non-iid data.
\newblock In \emph{Proceedings of the IEEE/CVF Conference on Computer Vision and Pattern Recognition (CVPR)}, pages 22841--22850, 2024.

\bibitem[Lubana et~al.(2022)Lubana, Tang, Kawsar, Dick, and Mathur]{lubana2022orchestra}
Ekdeep Lubana, Chi~Ian Tang, Fahim Kawsar, Robert Dick, and Akhil Mathur.
\newblock Orchestra: Unsupervised federated learning via globally consistent clustering.
\newblock In \emph{Proceedings of the 39th International Conference on Machine Learning}, pages 14461--14484. PMLR, 2022.

\bibitem[McMahan et~al.(2017)McMahan, Moore, Ramage, Hampson, and y~Arcas]{mcmahan2017fedavg}
Brendan McMahan, Eider Moore, Daniel Ramage, Seth Hampson, and Blaise~Aguera y Arcas.
\newblock Communication-efficient learning of deep networks from decentralized data.
\newblock In \emph{Artificial intelligence and statistics}, pages 1273--1282. PMLR, 2017.

\bibitem[Rehman et~al.(2023)Rehman, Gao, De~Gusm{\~a}o, Alibeigi, Shen, and Lane]{rehman2023ldawa}
Yasar Abbas~Ur Rehman, Yan Gao, Pedro Porto~Buarque De~Gusm{\~a}o, Mina Alibeigi, Jiajun Shen, and Nicholas~D Lane.
\newblock L-dawa: Layer-wise divergence aware weight aggregation in federated self-supervised visual representation learning.
\newblock In \emph{Proceedings of the IEEE/CVF international conference on computer vision}, pages 16464--16473, 2023.

\bibitem[Robbins and Monro(1951)]{herbert1951sgd}
Herbert Robbins and Sutton Monro.
\newblock {A Stochastic Approximation Method}.
\newblock \emph{The Annals of Mathematical Statistics}, 22\penalty0 (3):\penalty0 400 -- 407, 1951.

\bibitem[Roy and Vetterli(2007)]{roy2007effrank}
Olivier Roy and Martin Vetterli.
\newblock The effective rank: A measure of effective dimensionality.
\newblock In \emph{2007 15th European signal processing conference}, pages 606--610. IEEE, 2007.

\bibitem[Shi et~al.(2023)Shi, Liang, Zhang, Tan, and Bai]{shi2023feddecorr}
Yujun Shi, Jian Liang, Wenqing Zhang, Vincent Tan, and Song Bai.
\newblock Towards understanding and mitigating dimensional collapse in heterogeneous federated learning.
\newblock In \emph{The Eleventh International Conference on Learning Representations}, 2023.

\bibitem[Vaswani et~al.(2017)Vaswani, Shazeer, Parmar, Uszkoreit, Jones, Gomez, Kaiser, and Polosukhin]{vaswani2017attention}
Ashish Vaswani, Noam Shazeer, Niki Parmar, Jakob Uszkoreit, Llion Jones, Aidan~N Gomez, {\L}ukasz Kaiser, and Illia Polosukhin.
\newblock Attention is all you need.
\newblock \emph{Advances in neural information processing systems}, 30, 2017.

\bibitem[Wang and Liu(2021)]{wang2021behavior}
Feng Wang and Huaping Liu.
\newblock Understanding the behaviour of contrastive loss.
\newblock In \emph{Proceedings of the IEEE/CVF conference on computer vision and pattern recognition}, pages 2495--2504, 2021.

\bibitem[Wang and Isola(2020)]{wang2020alignuniform}
Tongzhou Wang and Phillip Isola.
\newblock Understanding contrastive representation learning through alignment and uniformity on the hypersphere.
\newblock In \emph{ICML}, 2020.

\bibitem[Xue et~al.(2024)Xue, Gan, Ni, Joshi, and Mirzasoleiman]{xue2024projectionhead}
Yihao Xue, Eric Gan, Jiayi Ni, Siddharth Joshi, and Baharan Mirzasoleiman.
\newblock Investigating the benefits of projection head for representation learning.
\newblock In \emph{The Twelfth International Conference on Learning Representations}, 2024.

\bibitem[Zhang et~al.(2023{\natexlab{a}})Zhang, Kuang, Chen, You, Shen, Xiao, Zhang, Wu, Wu, Zhuang, et~al.]{zhang2023fedca}
Fengda Zhang, Kun Kuang, Long Chen, Zhaoyang You, Tao Shen, Jun Xiao, Yin Zhang, Chao Wu, Fei Wu, Yueting Zhuang, et~al.
\newblock Federated unsupervised representation learning.
\newblock \emph{Frontiers of Information Technology \& Electronic Engineering}, 24\penalty0 (8):\penalty0 1181--1193, 2023{\natexlab{a}}.

\bibitem[Zhang et~al.(2023{\natexlab{b}})Zhang, Zhu, Chen, Song, Koniusz, and King]{zhang2023gcf}
Yifei Zhang, Hao Zhu, Yankai Chen, Zixing Song, Piotr Koniusz, and Irwin King.
\newblock Mitigating the popularity bias of graph collaborative filtering: A dimensional collapse perspective.
\newblock In \emph{Thirty-seventh Conference on Neural Information Processing Systems}, 2023{\natexlab{b}}.

\bibitem[Zhao et~al.(2018)Zhao, Li, Lai, Suda, Civin, and Chandra]{zhao2018fedshare}
Yue Zhao, Meng Li, Liangzhen Lai, Naveen Suda, Damon Civin, and Vikas Chandra.
\newblock Federated learning with non-iid data.
\newblock \emph{arXiv preprint arXiv:1806.00582}, 2018.

\bibitem[Zhuang et~al.(2021)Zhuang, Gan, Wen, Zhang, and Yi]{zhuang2021fedu}
Weiming Zhuang, Xin Gan, Yonggang Wen, Shuai Zhang, and Shuai Yi.
\newblock Collaborative unsupervised visual representation learning from decentralized data.
\newblock In \emph{Proceedings of the IEEE/CVF international conference on computer vision}, pages 4912--4921, 2021.

\bibitem[Zhuang et~al.(2022)Zhuang, Wen, and Zhang]{zhuang2022fedema}
Weiming Zhuang, Yonggang Wen, and Shuai Zhang.
\newblock Divergence-aware federated self-supervised learning.
\newblock In \emph{International Conference on Learning Representations}, 2022.

\end{thebibliography}
}
\clearpage
\appendix
\section{Experimental Details}

\subsection{Training} 
\label{app:training}
We follow the image augmentations used in SimCLR~\cite{chen2020simclr} and adopt ResNet-18~\cite{he2016resnet}   as the encoder, coupled with a two-layer linear projector. The model is trained for 5 epochs over 100 communication rounds with a batch size of 128. Both the encoder and projector produce output representations of 512 dimensions. Optimization is performed using SGD~\cite{herbert1951sgd} for both local and global models, with a learning rate of 0.1. The hyperparameters $\beta, \gamma,  \delta$ are set to 1.0, 1.0, and 0.1, respectively. The scaling factor $\alpha$ is set to 10.

\subsection{Effective Rank} 
\label{app:eff_rank}
\begin{definition}[Effective Rank]
    Let  matrix $\boldsymbol{Z} \!\in \!\mathbb{R}^{N \times d}\!$ with  $\boldsymbol{Z}\!=\!\boldsymbol{U} \boldsymbol{\Sigma} \boldsymbol{V}^T\!$ as its singular value decomposition, where $\boldsymbol{\Sigma}$ is a diagonal matrix with singular values $\sigma_1\! \geq\!  \cdots\! \geq\!\sigma_Q\! \geq \!0$ with $Q\!=\!\min(N, d)$. The distribution of singular values is defined as the normalized form $p_i=\sigma_i / \sum_{k=1}^Q\left|\sigma_k\right|$. The effective rank of the matrix $\boldsymbol{Z}$, 
    is defined as 
    \begin{equation}
        \operatorname{ERank}(\boldsymbol{Z})=\exp \left(H\left(p_1, p_2, \cdots, p_Q\right)\right)
    \end{equation}
    where $H\left(p_1, p_2, \cdots, p_Q\right)$ is the Shannon entropy  $H\left(p_1, p_2, \cdots, p_Q\right)=-\sum_{k=1}^Q p_k \log p_k$.
\end{definition}

\section{Additional Experiments}

\vspace{3pt}

\noindent\textbf{Distillation methods.} We compare two projector distillation methods, MSE and KL divergence, as studied in ~\cite{kim2021klmse}. The results indicate that both methods achieve similar performance and consistently outperform the baseline.

\begin{table}[h]
  \centering
    \caption{\textbf{Distillation methods.} Both MSE and KL divergence for PD achieve comparable performance and uniformity.}
  \resizebox{\linewidth}{!}{
  \begin{tabular}{ccccc}
    \toprule
     & LP & FT 1\% & FT 10\% & $-\mathcal{L}_{\text{uniform}} (\uparrow)$  \\
    \midrule
FedAlignUniform~\cite{wang2020alignuniform}  & 80.84 & 69.99 & 81.00 &  3.79 \\ 
SSD (MSE) & \bf 81.88 & 70.61 & 81.62 & 3.83 \\
SSD (KL) & 81.32 & \bf 70.74 & \bf 81.67 & \bf 3.84 \\
    \bottomrule
  \end{tabular}
  }

  \label{tab:distillation}
\end{table}

\end{document}